\documentclass[lettersize,journal]{IEEEtran}
\usepackage{amsmath,amsfonts}
\usepackage{algorithmic}
\usepackage{array}
\usepackage{graphicx}
\usepackage[caption=false,font=normalsize,labelfont=sf,textfont=sf]{subfig}
\usepackage{textcomp}
\usepackage{stfloats}
\usepackage{url}
\usepackage{verbatim}
\def\BibTeX{{\rm B\kern-.05em{\sc i\kern-.025em b}\kern-.08em
    T\kern-.1667em\lower.7ex\hbox{E}\kern-.125emX}}
\usepackage{balance}

\begin{document}

\title{Uni-Removal: A Semi-Supervised Framework for Simultaneously Addressing Multiple Degradations in Real-World Images}

\author{Yongheng Zhang, Danfeng Yan$\dagger$, Yuanqiang Cai 
\thanks{Y. Zhang, D. Yan, and Y. Cai are with State Key Laboratory Of Networking And Switching Technology, Beijing University of Posts and Telecommunications, Beijing, 100876, China; with the School of Computer Science, Beijing University of Posts and Telecommunications, Beijing, 100876, China. (e-mail: zhangyongheng@bupt.edu.cn; yandf@bupt.edu.cn; caiyuanqiang@bupt.edu.cn).}

\thanks{$\dagger$ Corresponding author (D. Yan).}
}

\markboth{IEEE Transactions on Circuits and Systems for Video Technology}
{Zhang \MakeLowercase{\textit{et al.}}: Uni-Removal: A Semi-Supervised Framework for Simultaneously Addressing Multiple Degradations in Real-World Images}


\maketitle

\begin{abstract}
Removing multiple degradations, such as haze, rain, and blur, from real-world images poses a challenging and ill-posed problem.
Recently, unified models that can handle different degradations have been proposed and yield promising results. 
However, these approaches focus on synthetic images and experience a significant performance drop when applied to real-world images. 
In this paper, we introduce Uni-Removal, a two-stage semi-supervised framework for addressing the removal of multiple degradations in real-world images using a unified model and parameters. 
In the knowledge transfer stage, Uni-Removal leverages a supervised multi-teacher and student architecture in the knowledge transfer stage to facilitate learning from pre-trained teacher networks specialized in different degradation types. 
A multi-grained contrastive loss is introduced to enhance learning from feature and image spaces.
In the domain adaptation stage, unsupervised fine-tuning is performed by incorporating an adversarial discriminator on real-world images. 
The integration of an extended multi-grained contrastive loss and generative adversarial loss enables the adaptation of the student network from synthetic to real-world domains. 
Extensive experiments on real-world degraded datasets demonstrate the effectiveness of our proposed method. 
We compare our Uni-Removal framework with state-of-the-art supervised and unsupervised methods, showcasing its promising results in real-world image dehazing, deraining, and deblurring simultaneously.
\end{abstract}


\begin{figure}[!t]
\centering
  \includegraphics[scale=0.51]{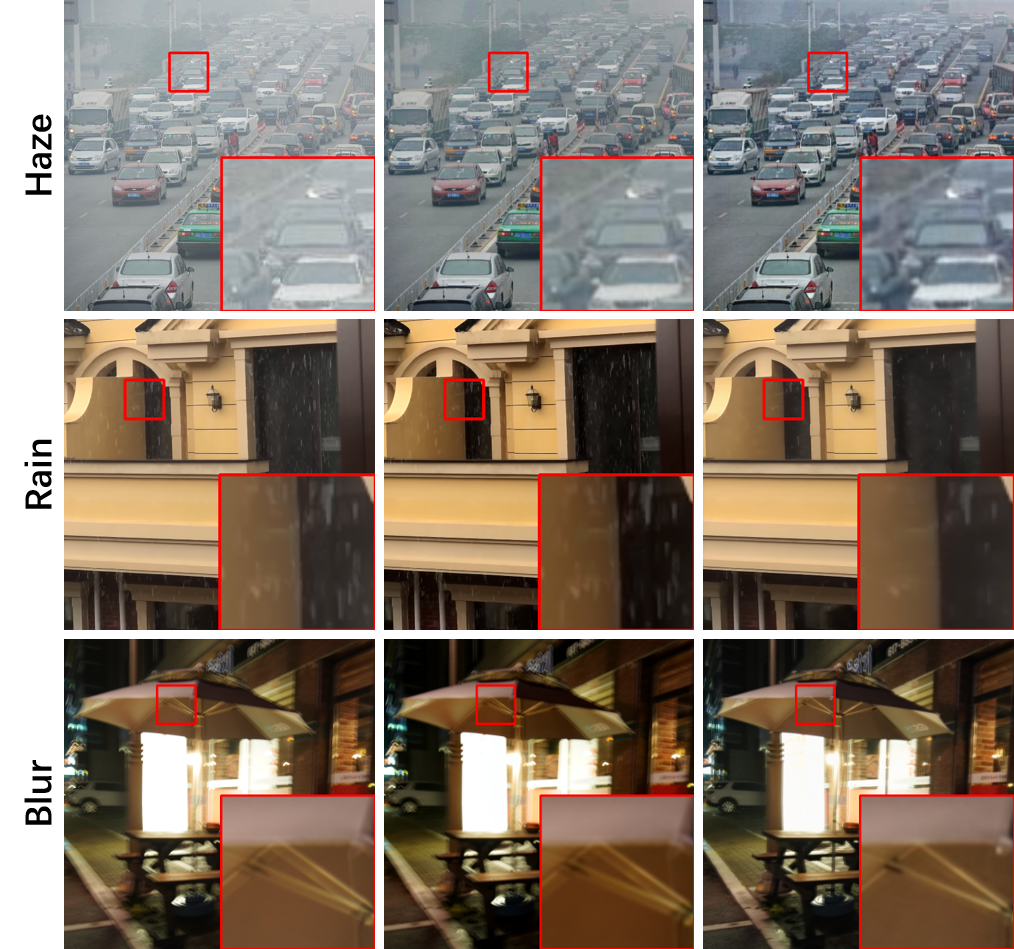}
  \begin{minipage}{0.1\linewidth}
  \centering
  \end{minipage}
  \begin{minipage}{0.3\linewidth}
  \centering
		(a) Degraded
  \end{minipage}
  \begin{minipage}{0.3\linewidth}
  \centering
		(b) MAWR \cite{chen2022learning}
  \end{minipage}
  \begin{minipage}{0.3\linewidth}
  \centering
		(c) Uni-Removal
  \end{minipage}
  \caption{Comparison of MAWR \cite{chen2022learning} and Uni-Removal on various real-world degraded images.
  (a), (b), and (c) depict real-world degraded images, followed by the results obtained from MAWR \cite{chen2022learning} and Uni-Removal, respectively. 
  Although MAWR \cite{chen2022learning} demonstrates competence in handling various degradations, its performance on real-world degraded images is limited.}
  
  \label{fig:firstpage_compare}
\end{figure}

\section{Introduction}
\IEEEPARstart{I}{mages} captured in the real world are susceptible to various degradations resulting from natural phenomena, limitations of shooting equipment, and transmission constraints, as depicted in Fig. \ref{fig:firstpage_compare} (a). These degradations not only impact the visual quality of the images but also hinder the accuracy of high-level visual tasks, including object detection, image segmentation, and text recognition. Consequently, the accurate and efficient removal of multiple degradations holds great significance for vision systems such as autonomous driving, security monitoring, and visual satellites.

Existing methods for degradation removal primarily focus on addressing individual degradations by leveraging handcrafted image priors \cite{tan2008visibility, he2010single, fattal2014dehazing,  hu2014deblurring, dong2011image, chen2015advanced}, such as the dark channel prior \cite{he2010single} and color-line prior \cite{fattal2014dehazing}. However, designing such priors is challenging, and their stability is often inadequate, leading to unsatisfactory outcomes. In the era of deep learning, some networks have been developed to estimate image prior parameters \cite{fu2017clearing, hyun2013dynamic, wang2021leveraging} as an initial step. Moreover, with the advancements in convolutional neural networks (CNNs) and transformers, several end-to-end models have emerged that directly predict clear images without degradations \cite{jonna2023distill, ren2018gated, zhang2021fish, song2022wsamf, liu2022multi, hao2022multi, cai2022multi, li2021dehazeflow, sharma2023wavelength}. 

While the aforementioned methods focus on removing specific degradations, there have been subsequent efforts to tackle the removal of multiple degradations using unified models \cite{zhang2017learning, zhang2019residual, zhang2020residual,  zamir2021multi, xie2021variational, mao2022deep}. 
Although these models exhibit good performance on various degradation removal tasks due to well-designed structures and selected functional units, they require switching between different sets of pre-trained parameters when tackling different types of degradations.

Recently, several All-in-One methods \cite{li2020all, valanarasu2022transweather, chen2022learning} have been proposed to address the removal of multiple degradations using a unified model and shared parameters. 
While some of these methods employ multiple encoders or degradation type embeddings for different degradations, they significantly reduce the number of parameters and demonstrate encouraging results. 
However, since these methods are trained on synthetically degraded images and corresponding clear images from real-world backgrounds, they often struggle to perform effectively in real-world scenes due to the substantial domain gap between degraded images in synthetic and real-world scenes.

\IEEEpubidadjcol

In this paper, we propose Uni-Removal, a two-stage semi-supervised framework for removing multiple degradations from real-world images, which can be applied to existing degradation removal networks. 
Uni-Removal aims to eliminate multiple degradations without incurring additional costs, relying solely on a unified model and a set of pre-trained parameters. 
The framework consists of a knowledge transfer stage and a domain adaptation stage, ensuring the model's ability to handle multiple degradations and improve its performance in real-world scenarios.

In the knowledge transfer stage, we initially train several teacher networks in a supervised manner using corresponding synthetic datasets. 
These teacher networks then guide the student network to acquire the knowledge of removing different types of degradations through a Multi-Grained Contrastive Learning (MGCL) loss, encompassing both feature-grained and image-grained aspects. 
The MGCL loss is calculated based on features and restored images obtained from both the teacher networks and the student network, using synthetic degraded images. 
Each feature and restored image from the teacher networks is treated as a positive example with respect to the corresponding feature and restored image from the student network, while other features and restored images from the student network within the same batch serve as negatives. 
This MGCL loss facilitates alignment between the student network and the teacher networks in both feature and image spaces.

In the domain adaptation stage, we further refine the student network by training it on unpaired real-world degraded and clear images, employing a discriminator in an adversarial setting. 
An EXtended Multi-Grained Contrastive Learning (EX-MGCL) loss is utilized in conjunction with the generative adversarial loss to fine-tune the student network's adaptation from the synthetic domain to the real-world domain. 
The EX-MGCL loss operates on features and restored images obtained from the student network using real-world degraded and clear images. 
A batch of features and restored images from real-world clear images are treated as positive examples with respect to feature and restored image from a real-world degraded image, while other features and restored images within the same batch of real-world degraded images serve as negatives. 
The EX-MGCL loss further facilitates the alignment of the student network with the real-world domain in both feature and image spaces.

These two types of contrastive losses contribute to faster learning of the student network at different levels and enhance the results on both synthetic and real-world datasets.

In summary, the contributions of this work are as follows:
\begin{itemize}
\item We propose Uni-Removal, a two-stage semi-supervised framework for removing multiple degradations from real-world images, utilizing a unified model and unified parameters.
This framework addresses the challenge of handling different types of real-world degradations while maintaining a consistent model architecture.

\item To enhance the performance of both the knowledge transfer stage and the domain adaptation stage, we introduce a Multi-Grained Contrastive Learning (MGCL) loss for the knowledge transfer stage and an Extended MGCL (EX-MGCL) loss for the domain adaptation stage. 
We conduct ablation studies to demonstrate the effectiveness of these loss functions.

\item Experimental results on real-world datasets demonstrate that Uni-Removal outperforms state-of-the-art all-in-one models, yielding promising results across various tasks, including real-world image dehazing, deraining, and deblurring.
Notably, on the SPA \cite{wang2019spatial} dataset, Uni-Removal shows significant improvements of 14.2 and 5.8 on BRISQUE \cite{mittal2012no} and PIQE \cite{venkatanath2015blind} quality assessment metrics, respectively, compared to the current state-of-the-art models.
\end{itemize}

\section{Related Work}
In this section, we provide a brief overview of the research related to single degradation removal methods, multiple degradations removal methods, knowledge distillation, and contrastive learning, which are relevant to our work.
\subsection{Single Degradation Removal}
\subsubsection{Supervised methods}

Over time, the performance of methods based on priors \cite{tan2008visibility, he2010single, fattal2014dehazing,  hu2014deblurring, dong2011image, chen2015advanced} become increasingly unsatisfactory.
The advent of deep convolutional neural networks (CNNs) and the availability of large-scale synthetic datasets have led to an increased interest in learning-based methods for single degradation removal. 
Early approaches \cite{ren2016single, cai2016dehazenet, zhang2018densely} focused on estimating parameters in physical scattering models using neural networks.

Subsequently, various end-to-end methods \cite{ren2018gated, li2021dehazeflow, wang2019spatial, zhang2021fish, song2022wsamf, liu2022multi, hao2022multi, cai2022multi, li2020dynamic} have been proposed to directly restore clear images without relying on explicit physical scattering models.
For instance, \textit{Zhang et al.} \cite{zhang2021fish} introduced a fish retina-inspired dehazing method that incorporates special retinal mechanisms to extract wavelength-dependent degradation information.  
\textit{Hao et al.} \cite{hao2022multi} addressed the challenge of size mismatch between rain streaks during the training and
testing phases by employing a monogenic wavelet transform-like hierarchy and a self-calibrated dual attention mechanism. 
\textit{Li et al.} \cite{li2020dynamic} incorporated depth information into CNN-based models for dynamic scene deblurring. 
Additionally, \textit{Li et al.} \cite{li2021dehazeflow} proposed DehazeFlow, the first work to utilize normalizing flows for single image dehazing.

\subsubsection{Semi-supervised and unsupervised methods}
Domain adaptation aims to bridge the gap between a source domain and a target domain.
Existing approaches \cite{long2015learning, bousmalis2017unsupervised, tsai2018learning} aim to align the source and target domains either at the feature level or pixel level by minimizing designed losses.

For instance, \textit{Kupyn et al.} \cite{kupyn2019deblurgan} proposed DeblurGAN-V2, a framework based on a relativistic conditional generative adversarial network (GAN) with a double-scale discriminator. 
They also introduced the feature pyramid network into deblurring. 
\textit{Shao et al.} \cite{shao2020domain} presented a domain adaptation framework that incorporates real hazy images into the training process using a cycle-GAN. Similarly, \textit{Wei et al.} \cite{wei2021deraincyclegan} employed a similar structure in image deraining and introduced a rain attention mechanism. 
\textit{Chen et al.} \cite{chen2021psd} explored a range of physical priors and developed a loss committee to guide the training on real hazy images.

While these methods demonstrate remarkable generalization performances for specific degradations, they often experience significant performance degradation when applied to other types of degradations.
In contrast, our proposed Uni-Removal framework effectively addresses the removal of multiple real-world degradations simultaneously.

\subsection{Multiple Degradations Removal}

\subsubsection{Multi-tasks methods}
Furthermore, several studies have explored the use of unified models to address multiple degradation removal problems \cite{ zhang2020residual, zamir2020learning, zamir2021multi, xie2021variational, mao2022deep}. 
For instance, \textit{Pan et al.} \cite{pan2018learning} proposed DualCNN, which incorporates two parallel branches to recover structures and details in an end-to-end manner. 
\textit{Zhang et al.} \cite{zhang2020residual} designed a residual dense network specifically for image restoration. 
\textit{Zamir et al.} \cite{zamir2021multi} adopted a multi-stage approach to restore degraded images and introduced an innovative per-pixel adaptive design that leverages in-situ supervised attention to reweight local features at each stage.
In addition, \textit{Mao et al.} \cite{mao2022deep} introduced an idempotent constraint into the deblurring framework, allowing the framework to be utilized for dehazing and deraining tasks as well.

These methods have demonstrated remarkable results across various degradation types using a unified framework. However, they typically require different sets of pre-trained weights for each type of degradation.

\subsubsection{All-in-One Degradations Removal}
Li et al. \cite{li2020all} proposed an end-to-end network with multiple encoders and a shared decoder, referred to as the All-in-One network. 
This network incorporates a discriminator to simultaneously assess the correctness and classify the degradation type of the enhanced images. 
Furthermore, an adversarial learning scheme is employed, wherein the loss of a specific degradation type is only backpropagated to the corresponding task-specific encoder.

Valanarasua et al. \cite{valanarasu2022transweather} developed a transformer-based end-to-end model consisting of a single encoder and a decoder. 
Specifically, the TransWeather model utilizes a novel transformer encoder with intra-patch transformer blocks to enhance attention within patches, along with a transformer decoder that incorporates learnable weather type embeddings to adapt to the specific weather degradation.

Chen et al. \cite{chen2022learning} adopted a two-stage knowledge learning process, which includes knowledge collation and knowledge examination, for adverse weather removal. 
In the collation stage, a collaborative knowledge transfer technique is proposed to guide the student model in integrating and learning the knowledge of various weather types from well-trained teacher models. 
In the examination stage, a multi-contrastive regularization approach is adopted to enhance the robustness of the student network for comprehensive weather removal.

Although these methods can handle different types of degradations using a single network, they often suffer a significant performance drop when applied to real-world degraded images due to the domain gap between synthetic and real-world degradations. 
To address this limitation, we propose Uni-Removal, a two-stage semi-supervised framework for multiple degradation removal in real-world images, aiming to adapt learning-based models to real degradation removal tasks.

\begin{figure*}[t]
\centering
  \includegraphics[scale=0.56]{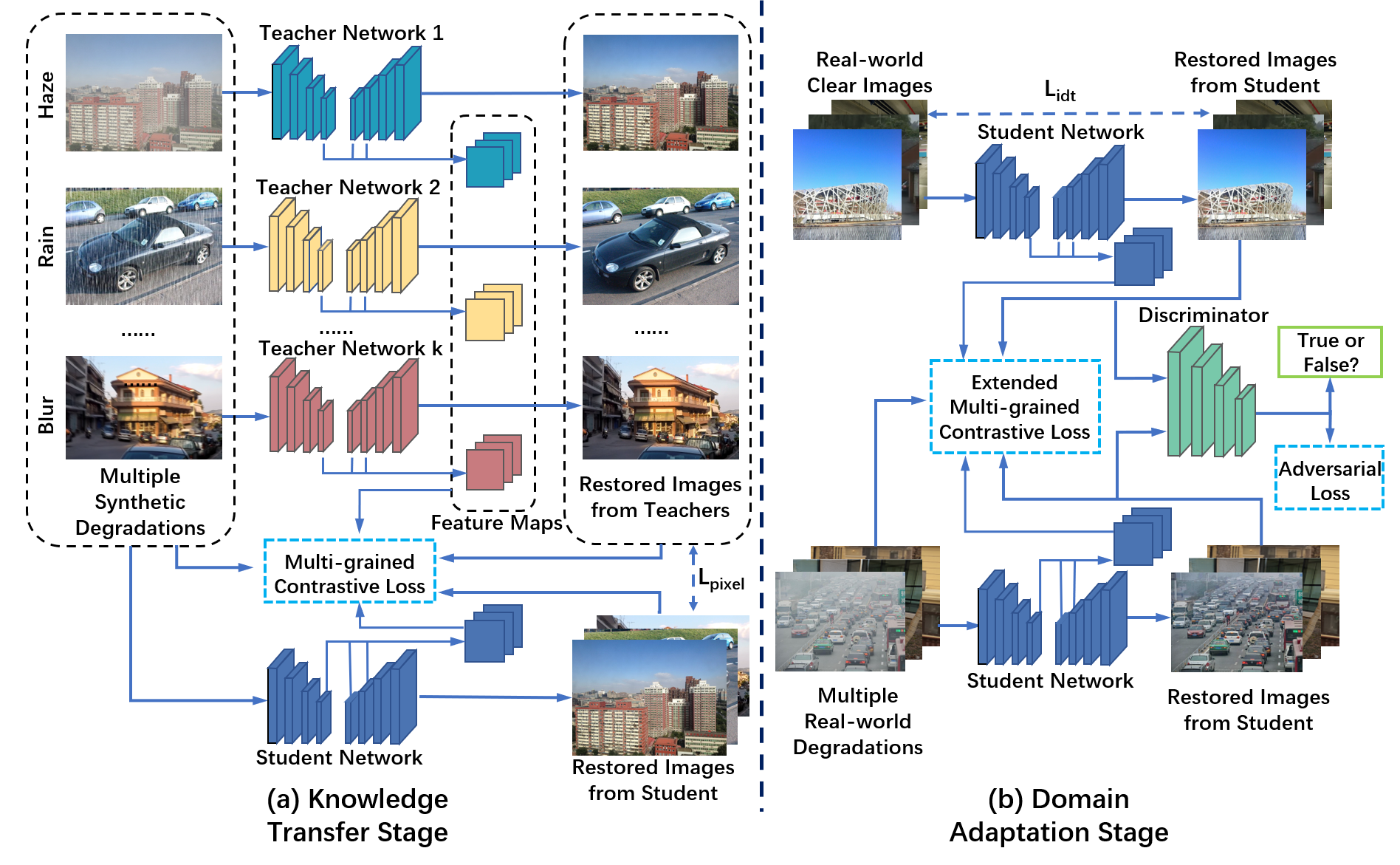}
  \caption{Framework of the proposed Uni-Removal. 
  In the (a) knowledge transfer stage, the student network is guided by multiple well-trained teacher networks, which have been trained on diverse synthetic degradation datasets. 
  The training process involves the utilization of a pixel-level L1 loss and a multi-grained contrastive learning loss.
  In the (b) domain adaptation stage, the student network undergoes unsupervised training using a discriminator on corresponding real-world degradation datasets. 
  In addition to the generative adversarial loss, the student network is further constrained by an identity loss and an extended multi-grained contrastive learning loss.}
  \label{fig:architecture}
\end{figure*}

\subsection{Knowledge Distillation}
Knowledge distillation \cite{hinton2015distilling} originally aimed to transfer knowledge from a large teacher model to a smaller student network. 
This involved training the student network on a transfer set while leveraging the soft target distribution provided by the larger model. 
However, it has been shown by \textit{Adriana et al.} \cite{adriana2015fitnets} that the teacher network does not necessarily have to be larger than the student network. 
In fact, both the outputs and intermediate representations learned by the teacher can enhance the training process and improve the final performance of the student network.

The concept of knowledge distillation has found wide application in various high-level computer vision tasks, including object detection \cite{wang2019distilling}, face recognition \cite{ge2018low}, and semantic segmentation \cite{liu2019structured}. 
More recently, researchers have also integrated knowledge distillation into image enhancement tasks \cite{hong2020distilling, chen2022learning}.

In contrast to traditional knowledge distillation methods that solely learn from positive examples provided by the teacher network, we propose a MGCL loss in the knowledge transfer stage to enable learning from both positive and negative examples.

\subsection{Contrastive Learning}
Contrastive learning is a technique that aims to sample positive and negative pairs from a given anchor point and then applies different contrastive losses to attract positive samples and repel negative samples \cite{gutmann2010noise, hermans2017defense, sohn2016improved, oord2018representation}.

In recent years, contrastive learning has been introduced into low-level vision tasks such as image-to-image translation \cite{park2020contrastive}, deraining \cite{Yuntong2022UnsupervisedDW, Chen2021UnpairedDI}, and dehazing \cite{wu2021contrastive, Chen2022UnpairedDI}. For instance, \textit{Chen et al.} \cite{Chen2022UnpairedDI} proposed an unsupervised contrastive CDD-GAN framework based on CycleGAN \cite{zhu2017unpaired} for image dehazing, where positive and negative samples are sampled from the hazy domain and clear domain, respectively. Similarly, \textit{Ye et al.} \cite{Yuntong2022UnsupervisedDW} devised a novel non-local contrastive learning mechanism that leverages the inherent self-similarity property for image deraining.

Building upon the concept of image-level contrastive learning, we extend the framework by introducing a comprehensive image-level and feature-level MGCL loss. Additionally, we incorporate an EX-MGCL loss that replaces one strongly correlated positive example with multiple weakly correlated positive examples.

\section{Proposed Method}
In this section, we present a normative definition of the task and provide an overview of our proposed method, including its underlying idea and overall structure. Additionally, we delve into the two training stages in detail and conclude with an introduction to the loss functions employed.

\subsection{Overview}
Our objective is to accurately restore corresponding clear images from different types of real-world degraded images without requiring changes to the methods or pre-trained models. 
Thus, it is essential to employ a method that can handle various degradation removal tasks simultaneously. 
However, training supervised methods solely on synthetic degraded datasets proves inadequate for effectively removing diverse degradations in real-world images due to the presence of domain shift. 
Consequently, fine-tuning the model using unsupervised training methods on real-world degraded datasets becomes necessary.

Addressing the aforementioned task requirements, we propose Uni-Removal, a semi-supervised framework for multiple degradations removal. 
Uni-Removal consists of two training stages: the knowledge transfer stage and the domain adaptation stage. 
As illustrated in Figure \ref{fig:architecture}, during the knowledge transfer stage, multiple pre-trained teacher networks guide the student degradation removal network, enabling the student network to acquire the capability to remove different types of degradations. 
Subsequently, in the domain adaptation stage, an unsupervised adversarial generative learning method is employed to simultaneously train the student network and a discriminator on real-world degradation removal datasets. 
This stage aims to enhance the student network's ability to remove various degradations in real-world images.

Uni-Removal is trained on synthetic datasets defined as 
$ X_{S_{i}}=\left\{x_{s_{i}}\right\}_{s_{i}=1}^{N_{S_{i}}}$,  
$ Y_{S_{i}}=\left\{y_{s_{i}}\right\}_{s_{i}=1}^{N_{S_{i}}}$, $ i={1,2...k}$ and real-world degraded datasets defined as $ X_{R_{i}}=\left\{x_{r_{i}}\right\}_{r_{i}=1}^{N_{R_{i}}}, i={1,2...k}$, $ X_{C}=\left\{x_{c}\right\}_{c=1}^{N_{C}}$, where k denotes the number of degradation types, $x_{s_{i}}$ denotes synthetic degraded image, $y_{s_{i}}$ denotes corresponding ground truth, $x_{r_{i}}$ denotes real degraded image, $x_{c}$ denotes real clear image, $N_{S_{i}}$, $N_{R_{i}}$ and $N_{C}$ denote the number of the synthetic image pairs of degradation type $i$, real degraded images of degradation type $i$, and real clear images, respectively.

\subsection{Knowledge Transfer Stage}
Simultaneously removing multiple degradations directly from different kinds of synthetic degradation datasets poses a significant challenge for a network. 
In contrast, learning to remove a specific degradation from a single type of synthetic degradation dataset is comparatively easier. 

To address this, instead of directly training a unified multi-degradation removal network, we adopt a two-step approach. 
Firstly, we train effective teacher networks for each degradation under supervision. 
The structure of each teacher network remains the same, but the parameters differ for different kinds of degradations. 
Specifically, we utilize the MPR-Net \cite{zamir2021multi} as our backbone and denote the teacher networks as $G_{T_{i}}, i={1,2...k}$.

Once the training of the teacher networks is completed, we fix their parameters and proceed to train a student network, denoted as $G_{S}$, leveraging the intermediate features and restored images produced by the teacher networks. 
The student network shares the same structure as each teacher network.

The training of the student network relies on two components. 
First, we employ a pixel-level L1 loss between the restored image generated by the student network and the corresponding restored image produced by the teacher network. 
Second, we utilize a Multi-Grained Contrastive Learning (MGCL) loss that operates at both the feature and image levels.

\begin{figure*}[t]
\centering
  \includegraphics[scale=0.7]{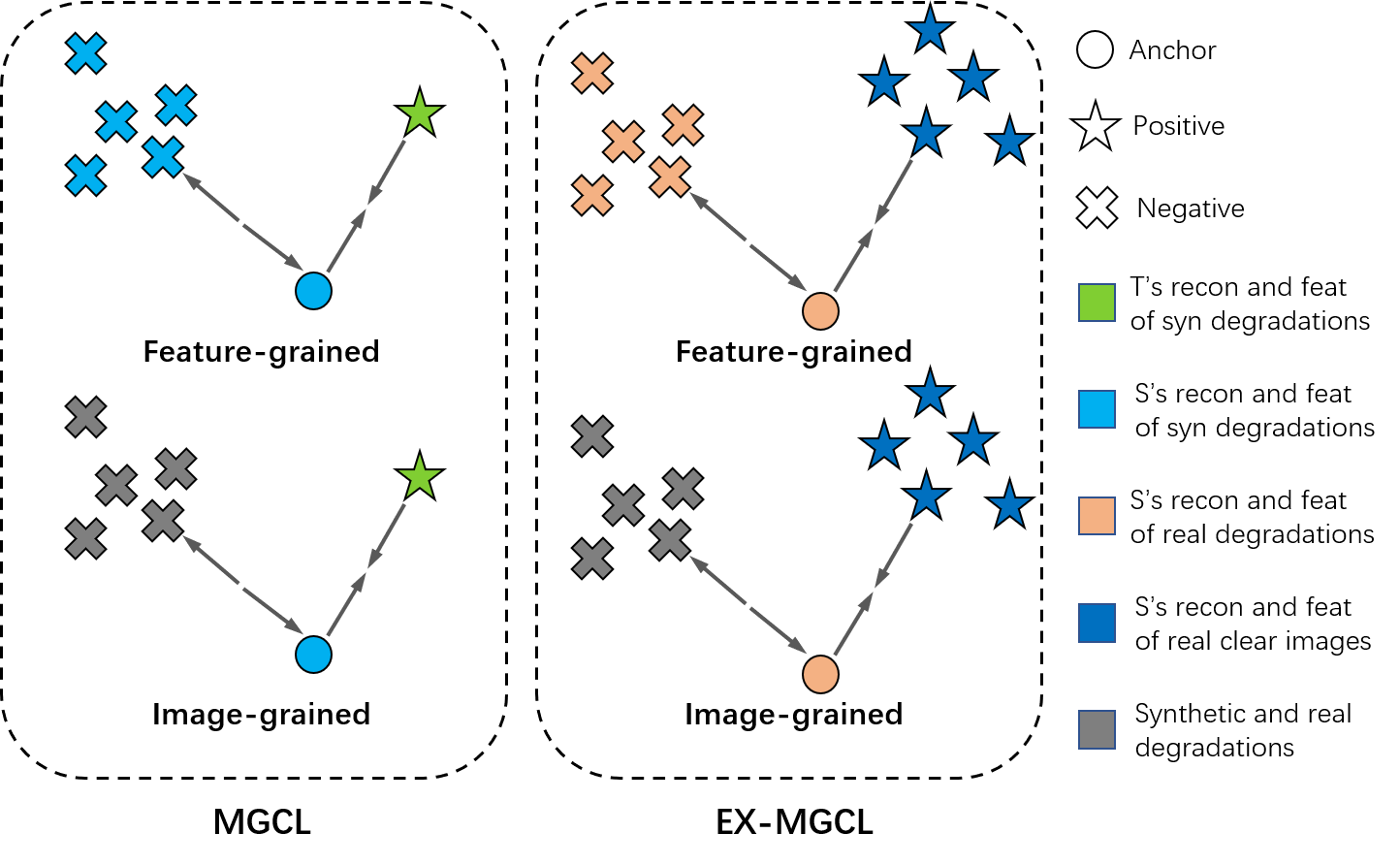}
  \caption{Interpretation of the proposed multi-grained contrastive learning (MGCL) loss and the extended multi-grained contrastive (EX-MGCL) learning loss. MGCL selects one  strongly correlated positive and multiple negatives for each sample. However, since a strongly correlated positive sample is not readily available in real-world degradation removal, the EX-MGCL replaces the strongly correlated positive with a batch of weakly correlated positives. This adaptation allows for a more effective learning process in the absence of strongly correlated positive samples in real-world scenarios.}
  \label{fig:MGCL}
\end{figure*}

As illustrated in Figure \ref{fig:MGCL}, the MGCL loss comprises a feature-grained contrastive loss and an image-grained contrastive loss. 
In the feature-grained contrastive loss, the intermediate features of the corresponding teacher network are treated as positives in relation to the intermediate features of the student network, while a batch of intermediate features from the student network with different degradation types are considered as negatives. 
In the image-grained contrastive loss, the restored image from the corresponding teacher network is regarded as positive with respect to the restored image from the student network, while a batch of synthetic images with various degradations serve as negatives.

The MGCL loss guides the student network to align with the teacher network both at the image level, by minimizing the distance to the corresponding teacher-restored image, and at the feature level, by minimizing the distance to the intermediate features of the corresponding teacher network. 
Simultaneously, it encourages the student network to be far away from synthetic degraded negative examples. 
To prevent misleading the fully trained student network, we assign a small trade-off weight to the feature-grained contrastive loss and gradually reduce it during training.

After the knowledge transfer stage, the student network gains the ability to remove different types of synthetic degradations. 
However, its performance in real-world image degradation removal remains unsatisfactory. 
Therefore, we proceed to further fine-tune the student network using real-world datasets in the domain adaptation stage.

\subsection{Domain Adaptation Stage}\label{chap:3.3}
In the domain adaptation stage, our objective is to enable the student network to effectively remove various degradations in real-world images while preserving the background unrelated to the degradations during the fine-tuning process.

To achieve this, the student network takes real-world degraded images and real-world clear images as input separately. 
The expected output is a set of restored images that are indistinguishable from real-world clear images. 
To accomplish this, we train a discriminator in an adversarial setting, where it aims to differentiate between the restored images and real-world clear images, while the student network strives to generate restored images that deceive the discriminator. 
This training is guided by a generating adversarial loss.

Furthermore, to ensure that real-world clear images remain unchanged before and after being processed by the student network, we incorporate an identity loss into the training process. 
This loss encourages the student network to preserve the essential details of the input clear images during restoration.

Additionally, we introduce an EXtend Multi-Grained Contrastive Learning (EX-MGCL) loss to facilitate the adaptation of the student network from the synthetic domain to the real-world domain. 
The EX-MGCL loss operates at both the feature and image levels, serving as a guiding principle for aligning the student network with the real-world clear domain.

As depicted in Figure \ref{fig:MGCL}, the EX-MGCL loss consists of a feature-grained contrastive loss and an image-grained contrastive loss. 
In the feature-grained contrastive loss, the intermediate features of the real-world degraded image take a batch of intermediate features of real-world clear images as positives, and take a batch of intermediate features of real-world degraded images of other degradation types as negatives. 
In the image-grained contrastive loss, the restored image of the real-world degraded image takes a batch of restored images of real-world clear images as positives, and takes a batch of real-world images with different degradations as negatives.

The EX-MGCL loss guides the student network to converge towards the real-world clear domain, both at the image level by minimizing its distance from the positive restored images and at the feature level by minimizing its distance from the positive intermediate features of real-world clear images. 
Simultaneously, it encourages the student network to move away from the real-world degraded domain. 
Similar to the knowledge transfer stage, we assign a small trade-off weight to the feature-grained contrastive loss and gradually decrease it during training to avoid misleading the fully trained student network.

Upon completion of the fine-tuning process in the domain adaptation stage, the student network possesses the capability to effectively remove various degradations in real-world images, utilizing a unified model with unified parameters.

\subsection{Training Losses}\label{chap:3.4}
The knowledge transfer stage and the domain adaptation stage are trained sequentially.

In the knowledge transfer stage, the student network $G_{S}$ is trained using a pixel-level loss $L_{pixel}$ and a MGCL loss $L_{m}$.

Given an image ${x_{s_{i}}}$ degraded by synthetic degradation $i$, the intermediate features and restored image of the corresponding teacher network are denoted as ${ft_{s_{i}}}$ and $ {xt_{s_{i}}}= G_{T_{i}}({x_{s_{i}}})$, respectively. 
The intermediate features and restored image of the student network are denoted as ${fs_{s_{i}}}$ and $ {xs_{s_{i}}} = G_{S}({x_{s_{i}}})$, respectively.
The pixel-level loss $L_{pixel}$ is defined as:

\begin{equation} 
L_{pixel} =\mathbb{E}_{x_{s_{i}} \sim X_{S_{i}}}[\|G_{S}({x_{s_{i}}})-G_{T_{i}}({x_{s_{i}}})\|_{1}], i=1,2...k.
\end{equation}

The MGCL loss includes a feature-grained contrastive loss and a image-grained contrastive loss. 
The common contrastive loss can be formulated as:

\begin{equation}
\begin{aligned}
&L_{C}(f, f^{+}, f^{-}) = \\
&-\log \frac{\operatorname{sim}\left(\phi(f), \phi(f^{+})\right)}{\operatorname{sim}\left(\phi(f), \phi(f^{+})\right)+\sum_{q=1}^{b} \operatorname{sim}\left(\phi(f), \phi(f_{q}^{-})\right)},
\end{aligned}
\end{equation}
where $f$, $f^{+}$, and$f_{j}^{-}$ denote the objective to be optimized, the positive sample, and the negative sample, respectively. $b$ denotes the number of negative samples, which is usually equal to the batch size. $\operatorname{sim}(u, v)=\exp \left(\frac{u^{T} v}{\|u\|\|v\| \tau}\right)$ denotes the similarity between two normalized feature vectors. $\tau$ denotes a scalar temperature parameter. $\phi()$ denotes a feature extraction operation by the VGG-19 \cite{simonyan2014very}, usually.

The feature-grained contrastive loss $L_{fg}$ and the image-grained contrastive loss $L_{ig}$ are defined as:
\begin{equation}
L_{fg} = L_{C}(fs_{s_{i}}, ft_{s_{i}}, \{\{fs^{q}_{s_{i}}\}^{k}_{i=1}\}^{b}_{q=1}), i=1,2...k,
\end{equation}
\begin{equation}
L_{ig} = L_{C}(xs_{s_{i}}, xt_{s_{i}}, \{\{x^{q}_{s_{i}}\}^{k}_{i=1}\}^{b}_{q=1}),i=1,2...k,
\end{equation}
where $fs_{s_{i}}$ and $xs_{s_{i}}$ are intermediate features and restored image from the student, the positive samples $ft_{s_{i}}$ and $xt_{s_{i}}$ are intermediate features and restored image from the teacher, the negative samples $\{\{fs^{q}_{s_{i}}\}^{k}_{i=1}\}^{b}_{q=1}$ and $\{\{x^{q}_{s_{i}}\}^{k}_{i=1}\}^{b}_{q=1}$ are a batch of intermediate features from the student and a batch of synthetic degraded images, respectively.

The MGCL loss and the overall loss of the knowledge transfer stage can be formulated as:
\begin{equation}
    L_{m} = L_{ig} + \alpha_{1}L_{fg},
\end{equation}
\begin{equation}
    L_{kt} = L_{pixel} + \alpha_{2}L_{m},
\end{equation}
where $\alpha_{1}$ and $\alpha_{2}$ are trade-off weights.

In the domain adaptation stage, the student network $G_{S}$ and the discriminator $D_{S}$ are trained using an adversarial loss $L_{gan}$, an identity mapping loss \cite{zhu2017unpaired} $L_{idt}$ and an EX-MGCL loss $L_{em}$.

Given an image ${x_{r_{i}}}$ degraded by real-world degradation $i$, the intermediate features and restored image of the student network are denoted as ${fs_{r_{i}}}$ and ${xs_{r_{i}}} = G_{S}({x_{r_{i}}})$, respectively.
For a clear real-world image $x_{c}$, the intermediate features and restored image of the student network are denoted as ${fs_{c}}$ and ${xs_{c}} = G_{S}({x_{c}})$, respectively.
Then the generate adversarial loss $L_{gan}$ can be formulated as:
\begin{equation}
\begin{aligned}
L_{gan} &=\mathbb{E}_{x_{r_{i}} \sim X_{R_{i}}} \left[D_{S}\left(G_{S}\left(x_{r_{i}}\right)\right)\right]\quad \\ &+ \mathbb{E}_{x_{c} \sim X_{C}}\left[D_{S}\left(G_{S}\left(x_{c}\right)\right)-1\right].
\end{aligned}
\end{equation}

The identity mapping loss is adopted to encourage the student network $G_{S}$ to preserve content information between the input real-world degraded images and output restored images. $L_{idt}$ can be defined as:
\begin{equation} 
L_{idt} =\mathbb{E}_{x_{c} \sim X_{C} }\left[\left\|G_{S}\left(x_{c}\right)-x_{c}\right\|_{1}\right].
\end{equation}

The EX-MGCL loss also includes a feature-grained contrastive loss and a image-grained contrastive loss.
However, since a strongly correlated positive sample is not available in real-world degradation removal, the extended contrastive loss is used, which replaces the strongly correlated positive with a batch of weakly correlated positives. 
The extended contrastive loss can be formulated as:
\begin{equation}
\begin{aligned}
&L_{EC}(f, f^{+}, f^{-}) = \\
&-\log \frac{\sum_{p=1}^{b}\operatorname{sim}\left(\phi(f),\phi( f_{p}^{+})\right)}{\sum_{p=1}^{b}\operatorname{sim}\left(\phi(f), \phi(f_{p}^{+})\right)+\sum_{q=1}^{b} \operatorname{sim}\left(\phi(f), \phi(f_{q}^{-})\right)},
\end{aligned}
\end{equation}

The extended feature-grained contrastive loss $L_{efg}$ and the extended image-grained contrastive loss $L_{eig}$ are defined as:
\begin{equation}
L_{efg} = L_{EC}({fs_{r_{i}}}, \{fs^{p}_{c}\}^{b}_{p=1}, \{\{fs^{q}_{r_{i}}\}^{k}_{i=1}\}^b_{q=1}),i=1,2...k,
\end{equation}
\begin{equation}
L_{eig} = L_{EC}(xs_{r_{i}}, \{xs^{p}_{c}\}^{b}_{p=1}, \{\{x^{q}_{r_{i}}\}^{k}_{i=1}\}^b_{q=1}),i=1,2...k,
\end{equation}
where $fs_{r_{i}}$ and $xs_{r_{i}}$ are intermediate features and restored image of the real-world degraded image, the positive samples $\{fs^{p}_{c}\}^{b}_{p=1}$ and $\{xs^{p}_{c}\}^{b}_{p=1}$ are intermediate features and restored images of a batch of real-world clear images, the negative samples $\{\{fs^{q}_{r_{i}}\}^{k}_{i=1}\}^b_{q=1}$ and $\{\{x^{q}_{r_{i}}\}^{k}_{i=1}\}^b_{q=1}$ are a batch of intermediate features of the real-world degraded images and a batch of real-world degraded images, respectively.

The EX-MGCL loss and the overall loss of the domain adaptation stage can be formulated as:
\begin{equation}
    L_{em} = L_{eig} + \lambda_{1}L_{efg},
\end{equation}
\begin{equation}
    L_{da} = L_{gan} + \lambda_{2}L_{idt} + \lambda_{3}L_{em},
\end{equation}
where $\lambda_{1}$, $\lambda_{2}$ and $\lambda_{3}$ are trade-off weights.

\section{Experiments}
In this section, we present the experimental setup and results to evaluate the effectiveness of our proposed framework. 
We implement our framework based on MPR-Net \cite{zamir2021multi} and conduct experiments on both synthetic and real-world degradation removal datasets. 
We compare our method against several state-of-the-art approaches and evaluate the visual quality and performance using commonly adopted metrics. 
Additionally, we perform two ablation studies to demonstrate the effectiveness of our proposed loss function and framework on both synthetic and real-world degradation removal datasets.

\begin{figure*}[htbp]
\centering
  \includegraphics[scale=0.545]{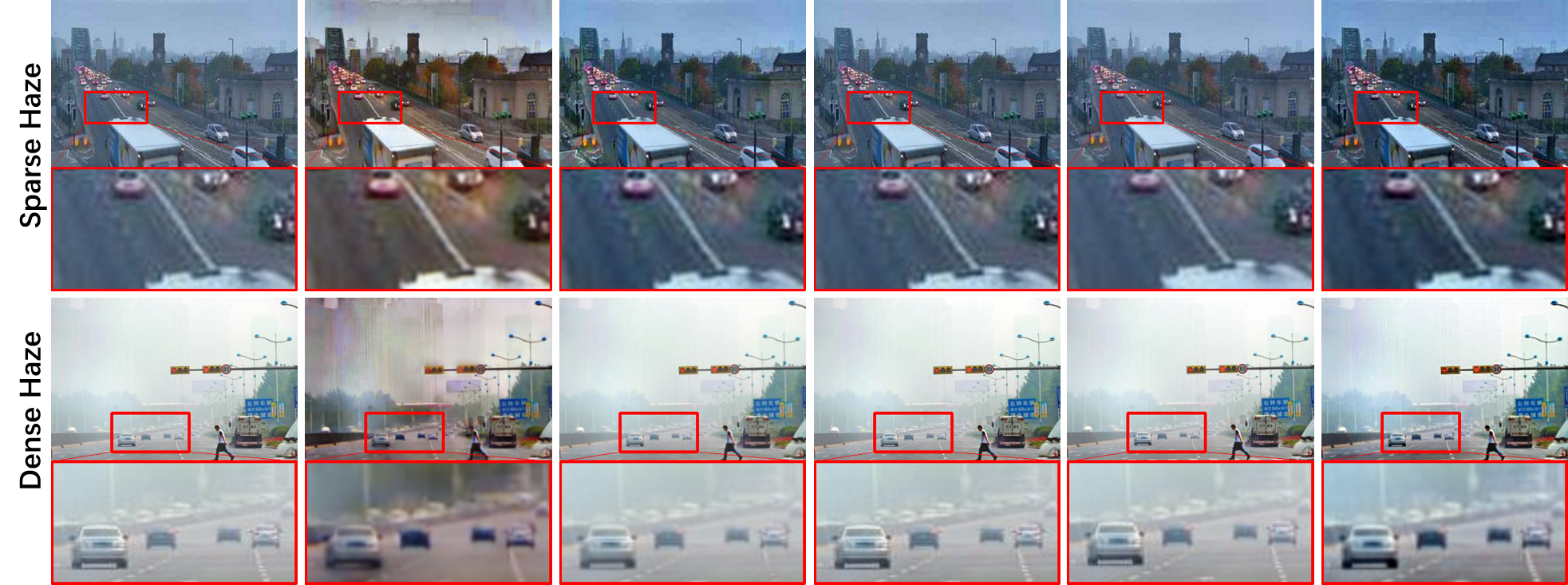}
  \begin{minipage}{0.1\textwidth}
  \centering
  \end{minipage}
  \begin{minipage}{0.18\textwidth}
  \centering
		(a) Hazy Images
  \end{minipage}
  \begin{minipage}{0.15\textwidth}
  \centering
		(b) DA-Net \cite{shao2020domain}
  \end{minipage}
  \begin{minipage}{0.15\textwidth}
  \centering
		(c) MAWR \cite{chen2022learning}
  \end{minipage}
  \begin{minipage}{0.155\textwidth}
  \centering
		(d) MPRNET \cite{zamir2021multi}
  \end{minipage}
  \begin{minipage}{0.17\textwidth}
  \centering
		(e) DehazeFlow \cite{li2021dehazeflow}
  \end{minipage}
  \begin{minipage}{0.145\textwidth}
  \centering
		(f) Uni-Removal
  \end{minipage}
    \caption{Comparison of dehazing results on real-world hazy images from RTTS \cite{li2018benchmarking}. 
    Uni-Removal demonstrates superior performance in effectively removing both sparse and dense haze while avoiding the introduction of artifacts, surpassing other existing methods.}
  \label{fig:RTTS results}
\end{figure*}

\begin{figure*}[htbp]
\centering
  \includegraphics[scale=0.566]{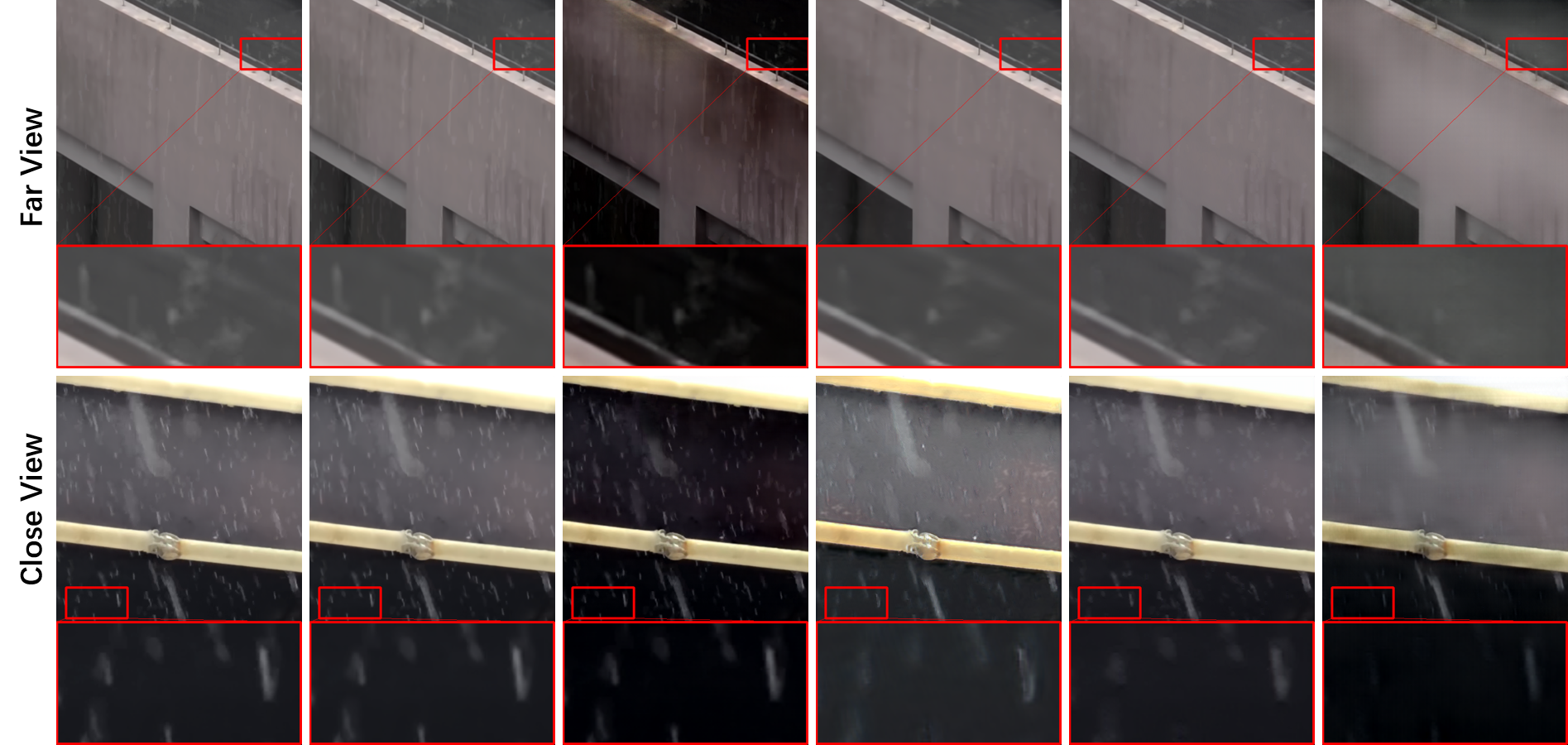}
  \begin{minipage}{0.1\textwidth}
  \centering
  \end{minipage}
  \begin{minipage}{0.185\textwidth}
  \centering
		(a) Rainy Images
  \end{minipage}
  \begin{minipage}{0.15\textwidth}
  \centering
		(b) MPR-Net \cite{zamir2021multi}
  \end{minipage}
  \begin{minipage}{0.145\textwidth}
  \centering
		(c) MAWR \cite{chen2022learning}
  \end{minipage}
  \begin{minipage}{0.145\textwidth}
  \raggedleft
		(d) NLCL \cite{Yuntong2022UnsupervisedDW}
  \end{minipage}
  \begin{minipage}{0.2\textwidth}
  \centering
		(e) DerainCycleGan \cite{wei2021deraincyclegan}
  \end{minipage}
  \begin{minipage}{0.135\textwidth}
  \centering
		(f) Uni-Removal
  \end{minipage}
    \caption{Comparison of deraining results on real-world rainy images from SPA \cite{wang2019spatial}. 
    Uni-Removal consistently achieves remarkable rain streak removal performance in both far view and close view scenes.
    It outperforms all other state-of-the-art methods, effectively removing the majority of rain streaks with exceptional quality.}
  \label{fig:SPA results}
\end{figure*}

\subsection{Implementation Details}
\subsubsection{Datasets}
For single image dehazing, we train and evaluate our method on the RESIDE \cite{li2018benchmarking} dataset. The RESIDE dataset consists of five subsets: Indoor Training Set (ITS), Outdoor Training Set (OTS), Synthetic Object Testing Set (SOTS), Unannotated real Hazy Images (URHI), and real Task-driven Testing Set (RTTS). 
We select the OTS and the URHI for training. 
For evaluation, we use the SOTS and the RTTS, which provides synthetic and real-world test images, respectively.

For single image deraining, we utilize the Rain1400 \cite{fu2017removing} dataset as our synthetic training and test set. 
This synthetic dataset contains 1400 image pairs with rain streaks of various sizes, shapes, and directions. 
To evaluate the generalization of our method, we fine-tune and test our model on the SPA dataset \cite{wang2019spatial}, which consists of real rainy images with diverse rain streak patterns.

For single image deblurring, we train and test our model on the GoPro \cite{nah2017deep} dataset. 
The synthetic GoPro dataset contains 2103 image pairs, with each pair consisting of a sharp image and a blurry image. 
We further fine-tune our model on the RealBlur \cite{rim2020real} dataset, which includes two subsets: RealBlur-J and RealBlur-R. 
The RealBlur-J subset consists of camera JPEG outputs, while the RealBlur-R subset is generated offline by applying white balance, demosaicking, and denoising operations to RAW images. 
Since the images in the RealBlur-J subset are more suitable for human observation, we only fine-tune and evaluate our method on this subset.

All subsequent experimental results are obtained based on the above datasets.
When training MAWR \cite{chen2022learning} and our Uni-Removal, we used a mixture of OTS \cite{li2018benchmarking}, Rain1400 \cite{fu2017removing}, and GoPro \cite{nah2017deep}, and when fine-tuning Uni-Removal, we used a mixture of URHI \cite{li2018benchmarking}, SPA \cite{wang2019spatial}, and BLUR-J \cite{rim2020real}.
In the training phase, all images are randomly cropped into patches of size $128 \times 128$. 
The pixel values of the patches are normalized to the range of -1 to 1.

\subsubsection{Training Details}
We implement our framework in PyTorch and utilize ADAM optimizer with a batch size of 16 to train the teacher and student networks on an Nvidia RTX3090. 
The temperature parameter $\tau$ = 1e-6 for both stages.
In the knowledge transfer stage, we train the student network for 400 epochs with the momentum $\beta_{1}$ = 0.9, $\beta_{2}$ = 0.999, and the learning rate is set as $2 \times 10^{-5}$. The trade-off weights are set as: $\alpha_{1}$ = 0.5 and $\alpha_{2}$ = 0.1.
There is a decay of 0.99 on $\alpha_{1}$ every epoch.
In the domain adaptation stage, we fine-tune the student network for 40 epochs. 
The momentum and the learning rate are set as: $\beta_{1}$ = 0.5, $\beta_{2}$ = 0.999, lr = $5 \times 10^{-6}$.  
The trade-off weights are set as: $\lambda_{1}$ = 0.5, $\lambda_{2}$ = 0.1, $\lambda_{3}$ = 0.1 and $\lambda_{4}$ = 0.01. 
There is also a decay of 0.99 on $\lambda_{1}$ every epoch.

\subsection{Comparisons with State-of-the-art Methods}
In this section, we compare our proposed Uni-Removal framework with several state-of-the-art methods on different types of real-world degradation removal datasets. 
We evaluate the performance both qualitatively and quantitatively, using commonly adopted metrics.
For the sake of fairness, we prefer to use the codes and the trained models provided by the authors.

\subsubsection{Visual Quality Comparison}

To evaluate the visual quality of Uni-Removal, we conducted experiments on the real-world haze removal dataset RTTS \cite{li2018benchmarking}, which is a subset of the RESIDE dataset \cite{li2018benchmarking}. 
We compared the results of Uni-Removal with four state-of-the-art dehazing methods: DA-NET \cite{shao2020domain}, MAWR \cite{chen2022learning}, MPR-Net \cite{zamir2021multi} (the backbone), and DehazeFlow \cite{li2021dehazeflow}. 
Both DehazeFlow \cite{li2021dehazeflow} and DA-NET \cite{shao2020domain} are networks designed for dehazing. DehazeFlow \cite{li2021dehazeflow} is a supervised dehazing network, while DA-NET \cite{shao2020domain} is semi-supervised. 
MPR-Net \cite{zamir2021multi} can remove different types of synthetic degradations using a unified model but different parameters. 
MAWR \cite{chen2022learning} is an "All-in-One" method for multiple synthetic degradations. 
The results are shown in Fig. \ref{fig:RTTS results}.

\begin{figure*}[htbp]
\centering
  \includegraphics[scale=0.6]{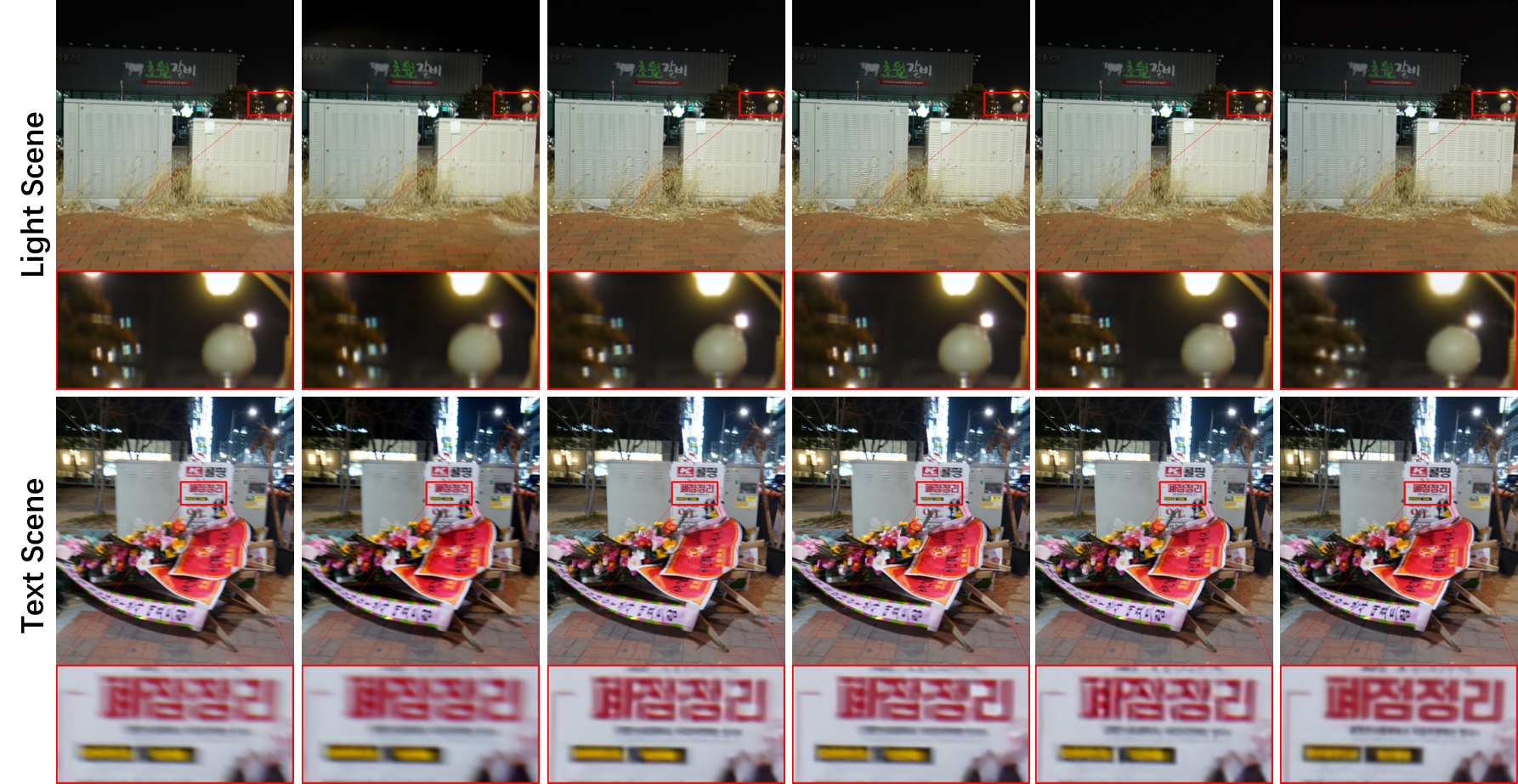}
  \begin{minipage}{0.1\textwidth}
  \centering
  \end{minipage}
  \begin{minipage}{0.195\textwidth}
  \centering
		(a) Blurry Images
  \end{minipage}
  \begin{minipage}{0.145\textwidth}
  \centering
		(b) MAWR \cite{chen2022learning}
  \end{minipage}
  \begin{minipage}{0.15\textwidth}
  \centering
		(c) XYDeblur \cite{ji2022xydeblur}
  \end{minipage}
  \begin{minipage}{0.145\textwidth}
  \raggedleft
		(d) MPR-Net \cite{zamir2021multi}
  \end{minipage}
  \begin{minipage}{0.18\textwidth}
  \centering
		(e) DeblurGan-v2 \cite{kupyn2019deblurgan}
  \end{minipage}
  \begin{minipage}{0.145\textwidth}
  \centering
		(f) Uni-Removal
  \end{minipage}
    \caption{Comparison of deblurring results on real-world blurry images from RealBlur-J \cite{rim2020real}.
    Uni-Removal demonstrates its capability to effectively mitigate the light ghosting artifact and deliver superior deblurring results, particularly in text scenes.
}
  \label{fig:BLUR-J results}
\end{figure*}

From the visual comparisons in Fig. \ref{fig:RTTS results}, it can be observed that DA-NET \cite{shao2020domain} removes a significant amount of haze, but the color of the images is shifted, and some black shadows appear. 
Methods that only perform supervised training on synthetic datasets, such as DehazeFlow \cite{li2021dehazeflow}, MPR-Net \cite{zamir2021multi}, or MAWR \cite{chen2022learning}, exhibit varying degrees of haze residue in their resulting images and lack the ability to effectively remove haze in real-world images.
In contrast, Uni-Removal generates images with the least haze residue and successfully preserves the background color, texture, and other details, achieving the best results in terms of visual quality.

Next, we evaluated the visual quality of Uni-Removal on the real-world rain streaks removal dataset SPA \cite{wang2019spatial}. 
We compared the results of Uni-Removal with four state-of-the-art deraining methods: MPR-Net \cite{zamir2021multi}, MAWR \cite{chen2022learning}, NLCL \cite{Yuntong2022UnsupervisedDW}, and DerainCycleGan \cite{wei2021deraincyclegan}. 
Both NLCL \cite{Yuntong2022UnsupervisedDW} and DerainCycleGan \cite{wei2021deraincyclegan} are unsupervised methods trained on real-world rain streaks removal datasets. 
The results are shown in Fig. \ref{fig:SPA results}.

As illustrated in Fig. \ref{fig:SPA results}, compared to MPR-Net \cite{zamir2021multi} trained on synthetic rainy datasets, unsupervised domain adaptation-based methods NLCL \cite{Yuntong2022UnsupervisedDW} and DerainCycleGan \cite{wei2021deraincyclegan} demonstrate better performance in effectively removing rain streaks from real-world images. 
Although MAWR \cite{chen2022learning} removes a majority of the rain streaks, the resulting images are generally too dark, leading to poor visual quality. 
In contrast, Uni-Removal significantly reduces the presence of rain streaks and successfully restores the background images with superior visual quality compared to the aforementioned methods.

Lastly, we assessed the visual quality of Uni-Removal on the real-world blur removal dataset RealBlur-J \cite{rim2020real}. We compared the results of Uni-Removal with four state-of-the-art deblurring methods: MAWR \cite{chen2022learning}, XYDeblur \cite{ji2022xydeblur}, MPR-Net \cite{zamir2021multi}, and DeblurGan-v2 \cite{kupyn2019deblurgan}. 
XYDeblur \cite{ji2022xydeblur} and DeblurGan-v2 \cite{kupyn2019deblurgan} are state-of-the-art supervised and unsupervised deblurring methods, respectively. 
The results are shown in Fig. \ref{fig:BLUR-J results}.

\begin{table}[tp]
\centering
  \caption{Quantitative results using NR-IQA metrics on RTTS \cite{li2018benchmarking}. The best results are in bold.}
  \label{tab:NR-IQA-RTTS}
  \begin{tabular}{cccl}
    \hline
    Method &    BRISQUE \cite{mittal2012no}$\downarrow$ & PIQE \cite{venkatanath2015blind}$\downarrow$ \\
    \hline
    Hazy &         37.011 & 51.254 \\
    DA-Net \cite{shao2020domain} &      32.456 & 50.787 \\
    MAWR \cite{chen2022learning} &  30.124  &  45.633 \\
    MPR-Net \cite{zamir2021multi} &          30.241 & 43.510 \\
    DehazeFlow \cite{li2021dehazeflow} &   26.059 & 38.879 \\
    \hline
    Uni-Removal(ours) &   \textbf{25.029} & \textbf{30.874} \\
    \hline
  \end{tabular}
\end{table}
 \begin{table}[tp]
\centering
  \caption{Quantitative results using NR-IQA metrics on SPA \cite{wang2019spatial}. The best results are in bold.}
  \label{tab:NR-IQA-SPA}
  \begin{tabular}{ccl}
    \hline
    Method &     BRISQUE \cite{mittal2012no}$\downarrow$ & PIQE \cite{venkatanath2015blind}$\downarrow$ \\
    \hline
    Rainy  & 68.025 & 44.388 \\
    MPR-Net \cite{zamir2021multi}  & 76.634 & 46.763 \\
    MAWR \cite{chen2022learning} &  54.721  &  44.652 \\
    NLCL \cite{Yuntong2022UnsupervisedDW}  & 56.952 & 43.286 \\
    DerainCycleGan \cite{wei2021deraincyclegan}  & 59.281 & 41.581 \\
    \hline
    Uni-Removal(ours)  & \textbf{40.489} & \textbf{35.786} \\
    \hline
  \end{tabular}
\end{table}

\begin{table}[tp]
\centering
  \caption{Quantitative results using NR-IQA metrics on RealBlur-J \cite{rim2020real}. The best results are in bold.}
  \label{tab:NR-IQA-BLUR}
  \begin{tabular}{ccl}
    \hline
    Method &     BRISQUE \cite{mittal2012no}$\downarrow$ & PIQE \cite{venkatanath2015blind}$\downarrow$ \\
    \hline
    Blurry  & 56.414 & 46.864 \\
    MAWR \cite{chen2022learning} &  44.919  &  43.636 \\
    XYDeblur \cite{ji2022xydeblur}  & 70.258 & 38.899 \\
    MPR-Net \cite{zamir2021multi}  & 64.643 & 36.165 \\
    DeblurGan-v2 \cite{kupyn2019deblurgan}  & 49.761 & \textbf{32.591} \\
    \hline
    Uni-Removal(ours)  & \textbf{39.966} & 34.517 \\
    \hline
  \end{tabular}
\end{table}

Similar to the results obtained in real-world image dehazing and deraining, the unsupervised deblurring method DeblurGan-v2 \cite{kupyn2019deblurgan} surpasses the supervised methods MAWR \cite{chen2022learning}, XYDeblur \cite{ji2022xydeblur}, and MPR-Net \cite{zamir2021multi}. 
Furthermore, Uni-Removal outperforms all the supervised blur removal methods and achieves comparable results to DeblurGan-v2 \cite{kupyn2019deblurgan} in terms of visual quality.

Overall, Uni-Removal demonstrates superior performance compared to the unified model MPR-Net \cite{zamir2021multi} and the all-in-one model MAWR \cite{chen2022learning} across various real-world degradation removal tasks. 
Furthermore, Uni-Removal exhibits promising results in terms of visual quality when compared to task-specific state-of-the-art semi-supervised and unsupervised methods.

\subsubsection{No-Reference Image Quality Assessment}
To further validate the effectiveness of the proposed framework, we conducted quantitative comparisons on real degradation datasets, namely RTTS \cite{li2018benchmarking}, SPA \cite{wang2019spatial}, and RealBlur-J \cite{rim2020real}.

Since paired real-world degraded images with clear backgrounds were not available as test sets, we could not adopt the commonly used reference evaluation indicators, such as PSNR and SSIM. 
Instead, we selected two general-purpose no-reference evaluation indicators: Blind/Referenceless Image Spatial Quality Evaluator (BRISQUE) \cite{mittal2012no} and Perception-based Image Quality Evaluator (PIQE) \cite{venkatanath2015blind} for quantitative comparisons.

BRISQUE utilizes a natural scene statistics model framework based on locally normalized luminance coefficients to quantify naturalness and quality in the presence of distortion. 
PIQE estimates quality from perceptually significant spatial regions, considering blockiness, blur, and noise. 
We calculated the results using the official MATLAB functions provided for these two evaluators. 
Lower values of these indicators indicate higher image quality. The results are summarized in Table \ref{tab:NR-IQA-RTTS}, Table \ref{tab:NR-IQA-SPA}, and Table \ref{tab:NR-IQA-BLUR}.

Table \ref{tab:NR-IQA-RTTS} demonstrates the superior performance of Uni-Removal in both indicators, surpassing other state-of-the-art methods by 1.03 and 8.01 on BRISQUE \cite{mittal2012no} and PIQE \cite{venkatanath2015blind}, respectively.  
These quantitative comparisons demonstrate that Uni-Removal effectively removes haze, reduces blur, and restores details in real-world dehazing scenarios. 

\begin{table*}
\centering
  \tabcolsep=15pt
  \caption{Ablation study of the MGCL loss in the knowledge transfer stage on three synthetic datasets. 'BKT' refers to base knowledge transfer training with only the pixel-level loss $L_{pixel}$. 'ICL' indicates training with image-grained contrastive learning loss, while 'FCL' denotes training with feature-grained contrastive learning loss.}
  \label{tab:ablation of kt}
  \begin{tabular}{ccccccl}
    \hline
    Dataset &    \multicolumn{2}{c}{SOTS\cite{li2018benchmarking}} & \multicolumn{2}{c}{Rain1400\cite{fu2017removing}} & \multicolumn{2}{c}{GoPro\cite{nah2017deep}} \\
    Method &  PSNR$\uparrow$  & SSIM$\uparrow$ & PSNR$\uparrow$  & SSIM$\uparrow$ & PSNR$\uparrow$  & SSIM$\uparrow$ \\
    \hline
    Teacher &     31.770 & 0.980 & 32.622 & 0.939 & 31.059 & 0.886 \\
    \hline
    BKT&    28.784 & 0.964 & 32.052 & 0.924 & 27.100 & 0.831 \\
    + ICL      &    28.387 & 0.968 & 32.141 & 0.926 & 28.445 & 0.868 \\
    + ICL + FCL &   \textbf{29.075} & \textbf{0.971} & \textbf{32.262} & \textbf{0.928} & \textbf{28.599} & \textbf{0.872} \\
    \hline
  \end{tabular}
\end{table*}

\begin{figure*}[htbp]
\centering
  \includegraphics[scale=0.731]{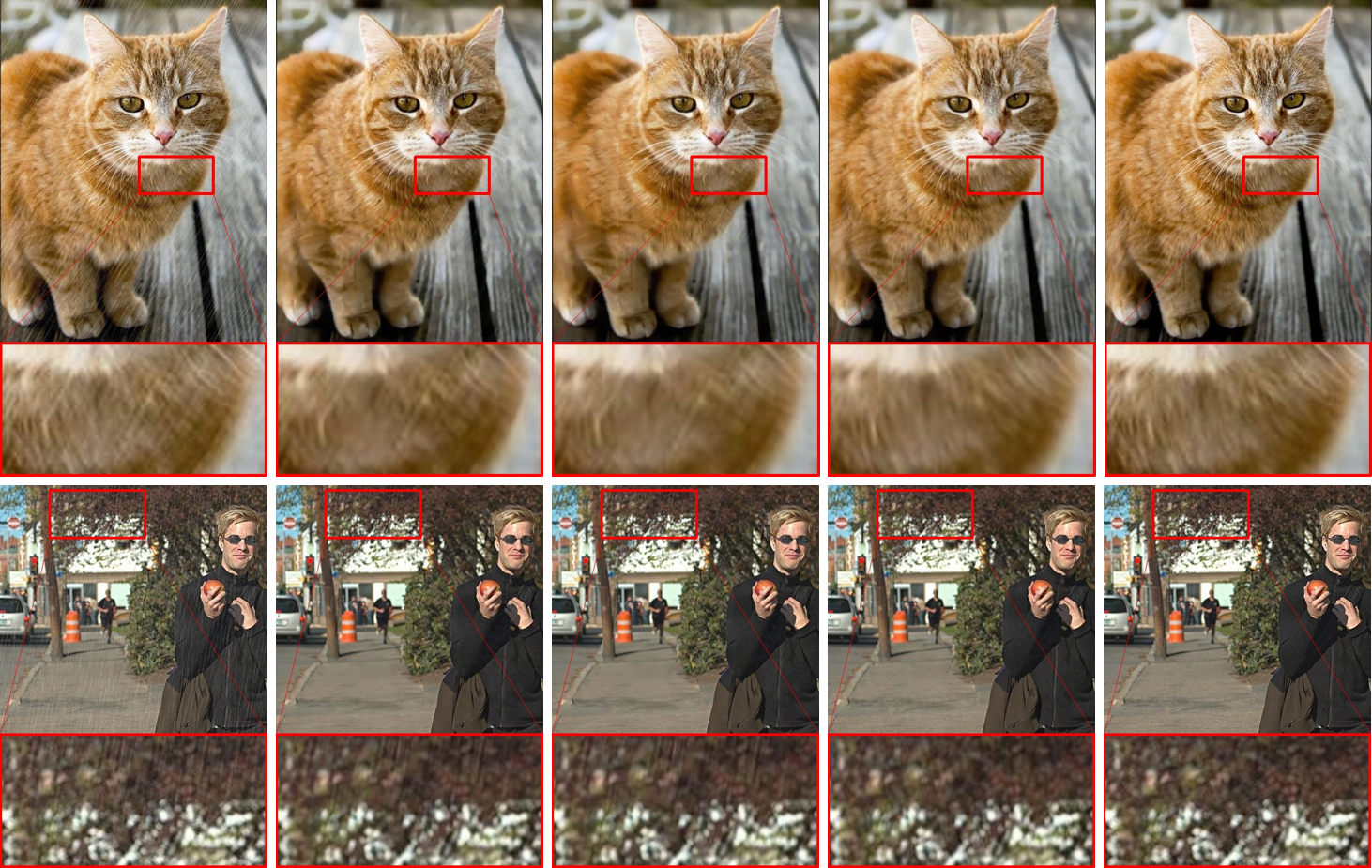}
  \begin{minipage}{0.19\textwidth}
  \centering
		(a) Rainy Images
  \end{minipage}
  \begin{minipage}{0.19\textwidth}
  \centering
		(b) BKT
  \end{minipage}
  \begin{minipage}{0.195\textwidth}
  \centering
		(c) +ICL
  \end{minipage}
  \begin{minipage}{0.195\textwidth}
  \centering
		(d) +ICL+FCL
  \end{minipage}
  \begin{minipage}{0.19\textwidth}
  \centering
		(e) GT
  \end{minipage}
    \caption{Ablation study of the MGCL loss in the knowledge transfer stage using synthetic rainy images. (a) and (e) represent the rainy images and their respective ground truth from Rain1400 \cite{fu2017removing}. 'BKT' refers to base knowledge transfer training with only the pixel-level loss $L_{pixel}$. 'ICL' indicates training with image-grained contrastive learning loss, while 'FCL' denotes training with feature-grained contrastive learning loss.}
  \label{fig:ablation study kt}
\end{figure*}

Moreover, Table \ref{tab:NR-IQA-SPA} showcases the superior performance of Uni-Removal in real-world image deraining. 
The substantial improvements of 14.2 on BRISQUE \cite{mittal2012no} and 5.8 on PIQE \cite{venkatanath2015blind} underscore the remarkable ability of Uni-Removal to address real-world deraining challenges, which aligns with the visual quality comparison results.

Additionally, Table \ref{tab:NR-IQA-BLUR} presents the promising results achieved by Uni-Removal in real-world deblurring tasks. 
The substantial improvement of 9.8 on BRISQUE \cite{mittal2012no} indicates that Uni-Removal effectively removes blur without introducing artifacts, further validating its performance in real-world scenarios.

The quantitative comparison results align with the visual quality assessment, further affirming the efficacy of Uni-Removal as a powerful framework for addressing multiple real-world degradations.

\begin{figure*}[htbp]
\centering
  \includegraphics[scale=0.622]{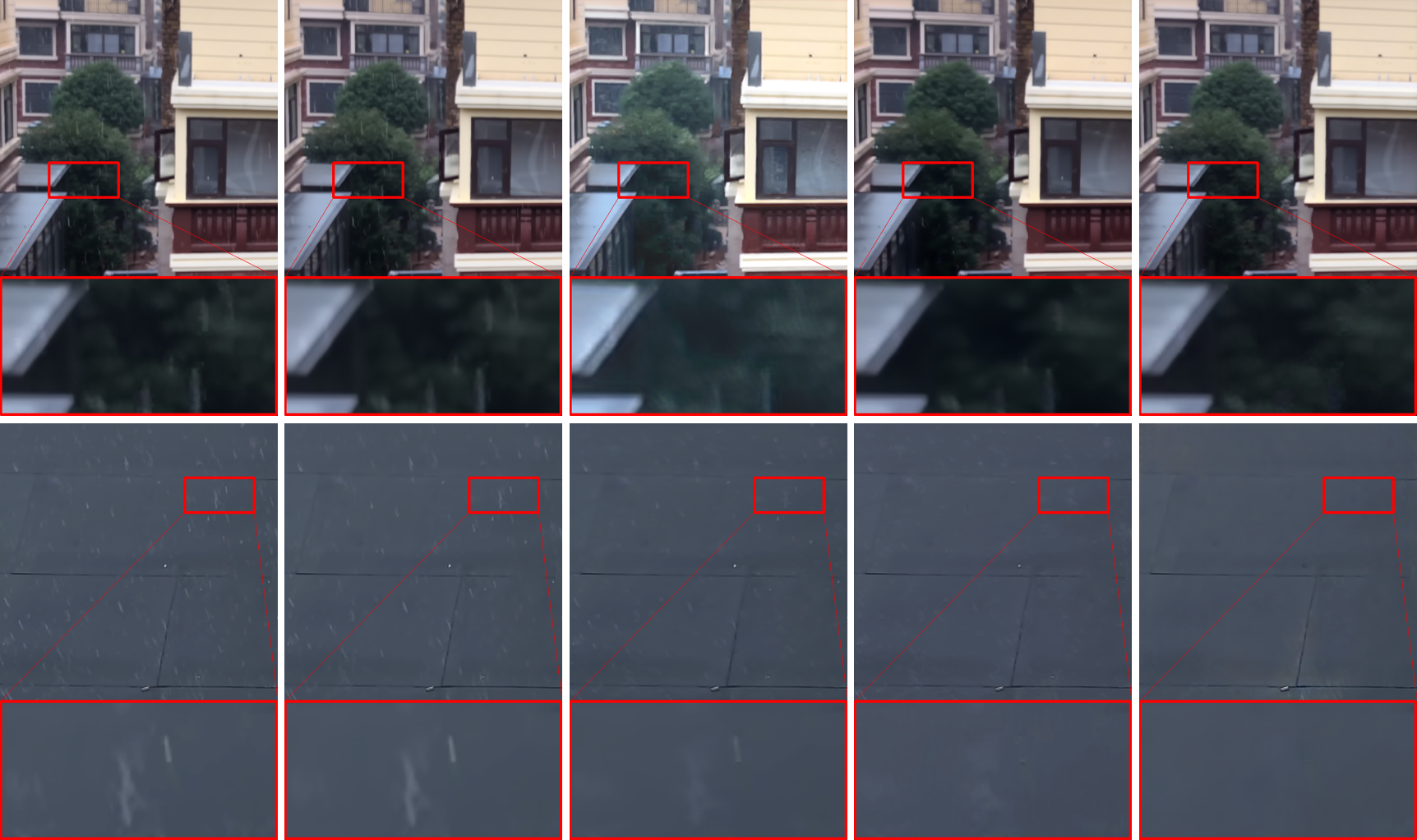}
  \begin{minipage}{0.195\textwidth}
  \centering
		(a) Rainy Images
  \end{minipage}
  \begin{minipage}{0.19\textwidth}
  \centering
		(b) KT
  \end{minipage}
  \begin{minipage}{0.19\textwidth}
  \centering
		(c) +BDA
  \end{minipage}
  \begin{minipage}{0.19\textwidth}
  \centering
		(d) +BDA+EICL
  \end{minipage}
  \begin{minipage}{0.21\textwidth}
  \centering
		(e) +BDA+EICL+EFCL
  \end{minipage}
    \caption{Ablation study of the domain adaptation stage and the EX-MGCL loss using real-world rainy images. (a) displays rainy images from SPA \cite{wang2019spatial}. 'KT' represents knowledge transfer with the MGCL loss. 'BDA' indicates base domain adaptation training with the identity mapping loss $L_{idt}$ and the adversarial loss $L_{gan}$. 'EICL' refers to training with the extended image-grained contrastive learning loss, while 'EFCL' denotes training with the extended feature-grained contrastive learning loss.}
  \label{fig:ablation study da}
\end{figure*}

\subsection{Ablation Study}
To validate the effectiveness of the proposed MGCL loss, EX-MGCL loss, and domain adaptation training strategy, we conducted ablation studies on synthetic degradation datasets and real-world degraded datasets.

Initially, we performed an ablation study on the knowledge transfer stage to demonstrate the effectiveness of the MGCL loss. 
We compared three combinations in the knowledge transfer stage using three synthetic degradation datasets: SOTS \cite{li2018benchmarking} (dehazing), Rain1400 \cite{fu2017removing} (deraining), and GoPro \cite{nah2017deep} (deblurring). 
We employed peak signal-to-noise ratio (PSNR) and structural similarity index (SSIM) as quantitative evaluation metrics. 
The results are presented in Table \ref{tab:ablation of kt}.

Table \ref{tab:ablation of kt} indicates that both the image-grained contrastive learning loss and the feature-grained contrastive learning loss improve the effectiveness of the knowledge transfer stage across all three datasets. 
The complete knowledge transfer model (BKT+ICL+FCL) outperforms the base knowledge transfer model by 0.291, 0.210, 1.499, and 0.007, 0.004, 0.041 in terms of PSNR and SSIM on the three synthetic datasets, respectively. 
Moreover, the complete knowledge transfer model shows minimal performance degradation compared to the teacher network trained on task-specific datasets.

Fig. \ref{fig:ablation study kt} provides visual results for the aforementioned combinations. 
The base knowledge transfer leaves residual rain streaks, which are further reduced by both the image-grained contrastive learning loss and the feature-grained contrastive learning loss. 
The background image restored by the complete knowledge transfer (BKT+ICL+FCL) not only exhibits the fewest residual rain streaks but also closely resembles the ground truth in terms of clarity and texture.

Next, we conducted an ablation study on the domain adaptation stage to verify the effectiveness of the domain adaptation training strategy and the EX-MGCL loss. 
We compared four combinations in the domain adaptation stage using three real-world degradation datasets: RTTS \cite{li2018benchmarking} (dehazing), SPA \cite{wang2019spatial} (deraining), and BLUR-J \cite{rim2020real} (deblurring). 
For quantitative comparison, we utilized BRISQUE \cite{mittal2012no} and PIQE \cite{venkatanath2015blind} as evaluation indicators. 
The results are presented in Table \ref{tab:ablation of da}.
\begin{table*}
\centering
  \caption{Ablation study of the domain adaptation stage and the EX-MGCL loss on three real-world datasets. 'KT' represents knowledge transfer with the MGCL loss. 'BDA' indicates base domain adaptation training with the identity mapping loss $L_{idt}$ and the adversarial loss $L_{gan}$. 'EICL' refers to training with the extended image-grained contrastive learning loss, while 'EFCL' denotes training with the extended feature-grained contrastive learning loss.}
  \label{tab:ablation of da}
  \begin{tabular}{ccccccl}
    \hline
    Dataset &    \multicolumn{2}{c}{RTTS\cite{li2018benchmarking}} & \multicolumn{2}{c}{SPA\cite{wang2019spatial}} & \multicolumn{2}{c}{BLUR-J\cite{rim2020real}} \\
    Method &  BRISQUE \cite{mittal2012no}$\downarrow$  & PIQE \cite{venkatanath2015blind}$\downarrow$ & BRISQUE \cite{mittal2012no}$\downarrow$  & PIQE \cite{venkatanath2015blind}$\downarrow$ & BRISQUE \cite{mittal2012no}$\downarrow$  & PIQE \cite{venkatanath2015blind}$\downarrow$ \\
    \hline
    Original degraded &     37.011 & 51.254 & 68.025 & 44.388 & 56.414 & 36.864 \\
    KT &     34.271 & 50.955 & 38.391 & 49.260 & 39.062 & 49.571 \\
    + BDA&  28.333   & 36.225 & 33.445 & 40.576 & 36.500 & 38.651 \\
    + BDA + EICL      &  25.961   & 31.152 & 32.057 & 38.874 & 35.478 & 36.196 \\
    + BDA + EICL + EFCL &   \textbf{24.029} & \textbf{28.874} & \textbf{30.689} & \textbf{35.786} & \textbf{34.966} & \textbf{34.517} \\
    \hline
  \end{tabular}
\end{table*}

As presented in Table \ref{tab:ablation of da}, the knowledge transfer model without domain adaptation exhibits poor performance on real-world degradation removal tasks. 
However, with domain adaptation, significant improvements are observed, with enhancements of 5.94, 4.95, 2.56 and 14.73, 8.69, 10.92 in terms of BRISQUE and PIQE on the three real-world datasets, respectively. 
Additionally, both the extended image-grained contrastive learning loss and the extended feature-grained contrastive learning loss contribute to the quality improvement of restored background images, including naturalness, color, distortion, and other factors. 
Compared to the base domain adaptation model, the full Uni-Removal (KT+BDA+EICL+EFCL) demonstrates further significant improvements of 5.94, 4.95, 2.56 and 14.73, 8.69, 10.92 in BRISQUE and PIQE on the three real-world datasets, respectively.

Fig. \ref{fig:ablation study da} illustrates the visual outcomes obtained from the four aforementioned combinations. 
The performance of the knowledge transfer model alone struggles when it comes to eliminating rain streaks in real-world images.
However, incorporating the domain adaptation training stage significantly enhances the model's ability to derain such images effectively. 
Moreover, the inclusion of the extended image-grained contrastive learning loss (EICL) and the extended feature-grained contrastive learning loss (EFCL) further contributes to improving the overall quality of the restored images. 
Notably, the background image restored using the complete Uni-Removal approach (KT+BDA+EICL+EFCL) exhibits an impressive outcome with minimal residue of rain streaks and no noticeable artifacts. 
These findings are supported by both quantitative and qualitative comparisons across the six synthetic and real-world datasets, providing compelling evidence for the effectiveness of the proposed MGCL loss, EX-MGCL loss, and domain adaptation training strategy.

\section{Conclusion}
In this paper, we introduced Uni-Removal, a two-stage semi-supervised framework designed for the removal of multiple degradations in real-world images. 
Our framework incorporates a multi-grained contrastive learning loss function, an extended multi-grained contrastive learning loss function, and a two-stage training strategy. 
Through the utilization of knowledge transfer and domain adaptation, Uni-Removal effectively addresses various degradations in real-world images, employing a unified model and unified parameters.
Extensive experiments conducted on synthetic and real-world datasets, covering dehazing, deraining, and deblurring tasks, demonstrate the effectiveness of our proposed loss functions, training strategy, and overall framework. 
The results highlight the capability of Uni-Removal to tackle multiple degradations, achieving significant improvements in terms of visual quality and quantitative evaluation metrics.
In future research, we plan to expand the evaluation to cover a wider range of real-world degradations and explore the applicability of Uni-Removal to other related tasks.
Overall, the proposed Uni-Removal framework presents a promising approach for addressing multiple degradations in real-world images, opening up possibilities for improved image restoration techniques in various practical applications.

\section*{Acknowledgments}
This work is supported by the National Key Research and Development Program of China with No. 2021YF3300700, Beijing University of Posts and Telecommunications Basic Research Fund with No. 2022RC12, and State Key Laboratory of Networking and Switching Technology with No. NST20220303.

\bibliographystyle{IEEEtran}
\bibliography{IEEEtrans}

\begin{IEEEbiography}
[{\includegraphics[width=1in,height=1.25in,clip,keepaspectratio]{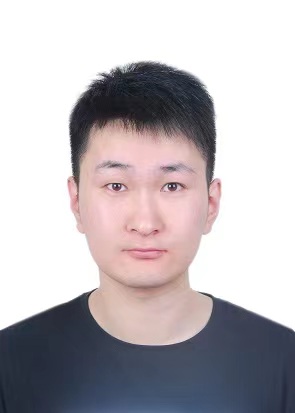}}]{Yongheng Zhang}
received the B.S. and M.S. degrees from the Beijing University of Posts and Telecommunications, Beijing, China, in 2017 and 2020, and he is currently pursuing a Ph.D. degree at the Beijing University of Posts and Telecommunications, Beijing. His research interests include computer vision and image enhancement.
\end{IEEEbiography}

\begin{IEEEbiography}
[{\includegraphics[width=1in,height=1.25in,clip,keepaspectratio]{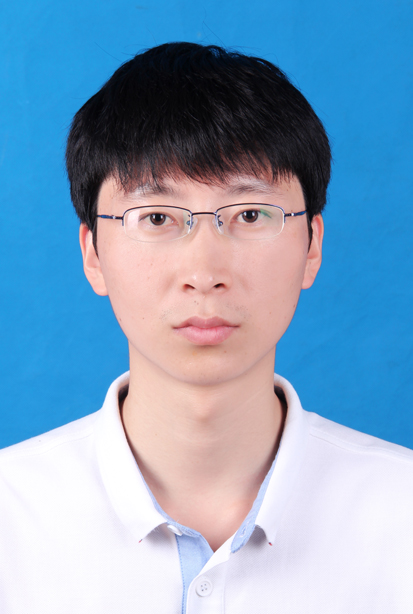}}]{Yuanqiang Cai}
received the Ph.D. degree with University of Chinese Academy of Sciences, Beijing, China, in 2021. He is currently a lecturer with the Beijing University of Posts and Telecommunications. His research interests include object detection, multimedia content analysis, and text localization and recognition in images and videos. He has published more than 10 papers in referred conference and journals including NeurIPS, AAAI, ACM MM, TCSVT, and PR.
\end{IEEEbiography}

\begin{IEEEbiography}
[{\includegraphics[width=1in,height=1.25in,clip,keepaspectratio]{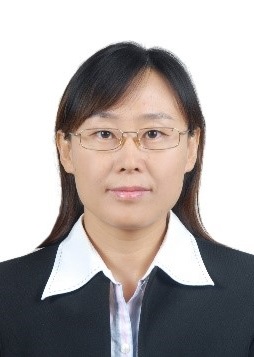}}]{Danfeng Yan}
is a professor and PhD supervisor at Beijing University of Posts and Telecommunications. She is working in State Key Laboratory of Network and Switching Technology, Beijing University of Posts and Telecommunications. She visited as a scholar at the School of Computer Science and Bell Labs, University of Ottawa, Canada in 2010. She has been engaged in scientific research and teaching of computer application technology for a long time. As the main researcher, She presides over or participates in the National High-tech R\&D Program (863 Program), National Basic Research Program (973 Program), National Key R\&D Program of China, Provincial and Ministerial Program, innovative research groups of the National Natural Science Foundation of China.
\end{IEEEbiography}

\vfill

\end{document}